\documentclass[12pt,reqno]{article}
\usepackage[english]{babel}
\usepackage{amsmath,dsfont, amsthm, amssymb}
\usepackage{amsfonts}
\usepackage{graphicx}
\usepackage{enumerate}
\usepackage[usenames,dvipsnames]{color}
\usepackage{subfigure}
\usepackage[right,pagewise,displaymath, mathlines]{lineno}
\usepackage{epstopdf}
\usepackage{color}
\usepackage{multirow}
\usepackage[table,xcdraw]{xcolor}
\usepackage[title]{appendix}

\usepackage{authblk}
\usepackage[round]{natbib}
\usepackage{float}

\usepackage{enumitem}

\usepackage{tikz}
\usepackage{schemabloc}

\usepackage[bottom]{footmisc}

\usepackage[colorlinks = true,
            linkcolor = blue,
            urlcolor = blue,
            citecolor = blue,
            breaklinks]{hyperref}

\usepackage{geometry}
\numberwithin{equation}{section}
\numberwithin{figure}{section}
\numberwithin{table}{section}

\theoremstyle{definition}

\numberwithin{equation}{section}

\definecolor{darkread}{rgb}{0.7, 0, 0}
\definecolor{darkbrown}{rgb}{0.55, 0.2, 0.15}
\definecolor{darkblue}{rgb}{0.1,0.1,0.6}
\definecolor{darkgreen}{rgb}{0.1,0.5,0.2}

\newcommand{\Fone}{\mathrm{F1}}

\newcommand{\TP}{\mathrm{TP}}
\newcommand{\TN}{\mathrm{TN}}
\newcommand{\FP}{\mathrm{FP}}
\newcommand{\FN}{\mathrm{FN}}
\newcommand{\MCC}{\mathrm{MCC}}
\newcommand{\Acc}{\mathrm{accuracy}}

\usepackage{setspace}
\usepackage[bottom]{footmisc}

\usepackage{tkz-graph}
\usepackage{pgfplots}

\title{\vspace{-15mm}\Large \bf Performance of diverse evaluation metrics in NLP-based assessment and text generation of consumer complaints \thefootnote\relax\footnotetext{
We are indebted to Jiandong Ren for his unwavering support and generous advice on the current version of the manuscript. This research has been supported by the NSERC, Canada Discovery Grant RGPIN-2022-04426.}
}

\author[,1]{\normalsize Peiheng Gao\thanks{Corresponding author; e-mail: \href{mailto:pgao47@uwo.ca}{pgao47@uwo.ca}}}

\author[,2,3]{Chen Yang\thanks{E-mail: \href{mailto:chen.yang@mountsinai.org}{chen.yang@mountsinai.org}}}

\author[,1]{Ning Sun\thanks{E-mail: \href{mailto:nsun46@uwo.ca}{nsun46@uwo.ca}}}

\author[,1]{Ri\v{c}ardas Zitikis\thanks{E-mail: \href{mailto:rzitikis@uwo.ca}{rzitikis@uwo.ca}}}

\affil[1]{\normalsize School of Mathematical and Statistical Sciences, Western University, London, Ontario~N6A~3K7, Canada}

\affil[2]{\normalsize Department of Population Health Science and Policy, Icahn School of Medicine at Mount Sinai, New York, New York~10029, USA}

\affil[3]{\normalsize Institute for Health Care Delivery Science, Icahn School of Medicine at Mount Sinai, New York, New York~10029, USA}

\begin{document}
\date{} 

\maketitle

\noindent {\small\vspace{-11mm}

\textbf{Abstract.}
    Machine learning (ML) has significantly advanced text classification by enabling automated understanding and categorization of complex, unstructured textual data. However, accurately capturing nuanced linguistic patterns and contextual variations inherent in natural language, particularly within consumer complaints, remains a challenge. This study addresses these issues by incorporating human-experience-trained algorithms that effectively recognize subtle semantic differences crucial for assessing consumer relief eligibility. Furthermore, we propose integrating synthetic data generation methods that utilize expert evaluations of generative adversarial networks and are refined through expert annotations. By combining expert-trained classifiers with high-quality synthetic data, our research seeks to significantly enhance machine learning classifier performance, reduce dataset acquisition costs, and improve overall evaluation metrics and robustness in text classification tasks.}

\medskip
{\it  Keywords and phrases}: Consumer complaints, Natural language processing, Human-experience-trained algorithms, Generative adversarial network.

\newpage
\section{Introduction}
ML is a discipline devoted on the development of algorithms and statistical models that enable computational systems to autonomously learn from data and perform tasks such as prediction, classification, and decision‐making without explicit programming. In text classification, ML techniques have dramatically enhanced the capability of automated systems to process and interpret large volumes of unstructured data. Complementarily, natural language processing (NLP) is an interdisciplinary field focused on the computational understanding, generation, and analysis of human language; its advancements empower classifiers to discern nuanced linguistic patterns and support applications ranging from sentiment analysis to topic categorization. Moreover, recent developments in text generation, such as generative adversarial networks (GANs), have facilitated the production of realistic synthetic textual data, further enriching training datasets and enhancing model robustness. Both ML and NLP approaches critically depend on the quality and diversity of training data to accurately capture the intricate structure of language, which frequently encompasses complexities such as slang, contextual meaning, and semantic variation.

Hence, incorporating human judgment into text classification processes is essential, as it facilitates the recognition of subtle linguistic variations and complex nuances inherent in natural language, particularly when assessing whether consumer complaints warrant relief. In \citet{Gao2025}, we present algorithms trained on human experience to analyze data obtained from \citet{CFPB}, which comprises a comprehensive collection of records detailing both monetary and non-monetary relief granted to consumer complaints. We contend that integrating human-experience-trained algorithms is critical for the effective performance of the ML classifiers.

While human-experience-trained algorithms are indispensable for ML classifiers (see, e.g., \cite{GZ2020b}) to capture subtle linguistic nuances in consumer complaints, the exclusive reliance on real-world data, which often exhibits various imperfections, underscores the necessity of integrating synthetic data generation into text classification pipelines. The effectiveness of ML classifiers is largely contingent upon the quality and representativeness of the training data. However, real world complaint datasets typically demonstrate significant variability in language usage, semantics, and contextual nuances.

Our research aims to enhance the effectiveness of ML-based text classifiers through two main strategies. First, we integrate the expertise embedded in expert-trained ML algorithms to facilitate robust comparisons among multiple classifiers during training. Second, we leverage expert annotations to generate high-quality synthetic complaint data that closely approximates real-world scenarios. Addressing these challenges is crucial not only for improving classification performance but also for reducing the costs and efforts associated with acquiring new complaint datasets, and this is one of the main purposes of our present study.

\subsection{Related studies}

The dataset utilized in this analysis, provided by \citet{CFPB}, consisted of textual records detailing a substantial number of monetary and non-monetary relief measures granted in response to consumer complaints \citep{HJP2021}. The consumer Financial Protection Bureau (CFPB) offered valuable insights, particularly when seeking to understand the underlying structure of consumer complaints. One of the major product categories is the mortgage complaints, \citet{DR2024} employed a simple linear regression model and conclude that the public disclosure of mortgage complaints enhances market discipline and improves financial protection for consumers. \citet{BP2021} examined regulatory challenges across different racial groups with respect to the quality of financial services using data from \citet{CFPB}. Their findings highlight an unintended adverse consequence of quantity-focused regulations on the quality of credit extended to lower-income and minority customers.

Selecting an appropriate model is crucial in the task of text classification, as it not only conserves computational resources but also enhances the quality of subsequent analyses. \citet{RA2022} fine-tuned the Text-to-Text Transfer Transformer \citep{RSRLNMZL2020} (T5) model for text summarization, effectively capturing key terms from the original text and subsequently improving classification performance. \citet{KLA2021} combined a convolutional neural network with a bidirectional long short-term memory network to develop a novel model that demonstrates excellent performance in classifying datasets, particularly narrative texts. \citet{HWY2023} deployed the FinBERT model, a variant of BERT fine-tuned on a financial textual dataset, which yield particularly strong performance on financial data.

Generated textual datasets are essential in research and practical applications, particularly when access to extensive, high quality real world data are limited by privacy concerns, regulatory constraints, or logistical challenges. By generating synthetic data that closely approximates structural and semantic complexities of natural language, researchers can perform controlled experiments to rigorously evaluate and refine natural language processing models under various conditions. \citet{Shahriar2022} offers an extensive survey of recent studies utilizing GANs for the generation of visual arts, music, and literary texts. In the context of literary text generation, the long short-term memory (LSTM) architecture has been widely adopted as the generator component within the GANs framework. Within the realm of deep generative models, \citet{IQ2022} compare the performance of several algorithms (e.g., recurrent neural networks (RNNs), LSTM, gated recurrent units, and bidirectional RNNs) and conclude that LSTM achieve the highest performance in generating textual data. \citet{IYKDS2022} extend traditional text generation architectures (e.g., RNN, LSTM, and convolutional neural networks) by introducing the CatGAN model; their comparative analysis on imbalanced datasets and datasets containing generated reviews demonstrates that CatGAN is capable of producing higher-quality text without sacrificing diversity.  \citet{LMLF2023} introduce a feature-aware conditional GAN (FA-GAN) for controllable, category-based text generation, concluding that FA-GAN not only meets the specified categorical requirements but also effectively captures the salient features of conditioned sentences, all while maintaining good readability, fluency, and authenticity.

In recent studies, generative adversarial networks are implement into various studies. \citet{XSWA2024} propose Market-GAN, a controllable generator with semantic context for financial simulation of the market features. 
\citet{WRZW2024} propose an encoder–decoder architecture to learn the latent variable distribution from a set of news–report data, thereby providing background knowledge. Next, authors employ a teacher–student network to distill knowledge, refining the output of the decoder component.
\citet{YHCWFD2024} present CTGGAN, a novel model for constraint-based text generation. The model uses a language model enhanced by a bias network to produce text that meets specified criteria. CTGGAN employs a discriminator that integrates sentiment and fluency evaluations to assess the quality of the generated text and guide improvements in the generator.
\citet{DPTDSM2025} integrate coalescent theory with a generative adversarial network to generate financial textual data and conclude that the resulting synthetic datasets are diverse and capture the complexity of authentic financial texts, thereby providing valuable resources for financial modeling and analysis.
\citet{KOIL2025} present a comprehensive survey of recent advances in the application of GANs to natural language processing and conclude that GANs can be effectively applied to textual data, achieving notable success in tasks such as text production, translation, and language modeling.

In evaluating model performance, researchers typically employ a variety of evaluation metrics. When addressing imbalanced datasets, \citet{Gao2025} utilize the $\Fone$ score as the primary metric. For binary classification tasks, \citet{CWJ2021} advocate the use of the Matthews correlation coefficient ($\MCC$), noting that it provides a more comprehensive assessment of classification performance. \citet{KK2022} underscore the benefits of employing Cohen’s kappa, particularly when assessing agreement levels in various natural language processing and text mining tasks.  \citet{KPKWH2024} apply Cohen’s kappa to compare the performance of large language models (LLMs) with that of human reviewers in systematic reviews, concluding that LLMs can rival human performance.

\subsection{Research gaps}
Several critical research gaps have been identified in the realm of textual data analysis within consumer complaint narratives. First, although NLP techniques have advanced substantially, most methodologies are designed for general-purpose applications and fail to capture the unique linguistic characteristics of consumer complaint narratives. These narratives often feature informal language, emotional expressions, colloquialisms, and industry-specific jargon related to various products and services. Also, consumer complaints frequently contain non-standard abbreviations, slang, and context-dependent expressions, which present significant challenges to generic NLP models, particularly in the absence of extensive domain-specific annotated corpora.

Second, inherent heterogeneity and unstructured nature pose significance of consumer complaint narratives challenges in data preprocessing, normalization, and information extraction. Since a diverse range of consumers provides these complaints, variations in writing style and vocabulary are common, particularly when addressing different products or services. This variability often results in inconsistencies (e.g., misspellings, divergent terminologies, and irregular syntactic structures) that complicate efforts to consolidate the data into a cohesive analytical framework. Consequently, traditional data processing pipelines, which are typically optimized for more homogeneous datasets, often struggle to effectively manage and analyze the multifaceted nature of consumer complaint narratives.

\subsection{Objective}
The primary objective of the present study is to analyze the different evaluation metrics that examine the impact of human biases on the NLP-based classification of consumer complaints provided by \citet{CFPB} into meritorious and non-meritorious categories. The investigation is structured around two principal steps:
\begin{enumerate}
    \item Evaluating the impact of human biases on various classification performance metrics, including accuracy, $\Fone$ score, Matthews correlation coefficient, and Cohen's kappa ($\mathrm{C}_{\kappa}$).  
    \item Assessing the performance of different classifiers on both raw narratives and generated narratives, employing a modified performance metric that accounts for  generative adversarial-based human-experience-trained algorithms \citep{Gao2025}.
\end{enumerate}

\subsection{Contributions}
The present research has several implications on how insurance companies, regulators, and others can better utilize massive volumes of customer complaints in order to train classifiers that support decision-making when resolving disputes. Among the achievements are:
\begin{enumerate}
    \item Modifying the traditional word embedding approach to leverage its advantages while reducing computational time.  
    \item Building a model that yields superior performance in detecting meritorious consumer complaints.
    \item Simulating a textual dataset that achieves performance comparable to that of the raw complaint narratives. 
    \item Demonstrating the utility of the proposed procedure for identifying meritorious customer complaints in the CFPB dataset through in-depth analysis.
    \end{enumerate}

The rest of the paper is organized as follows. 
In Section~\ref{data}, a subset of the CFPB consumer complaint data are chosen to illustrate the proposed procedure. 
In Section~\ref{Narrative_classfication}, we transform the raw narratives into quantitative representations using two novel featurization approaches and assess their performance with a diverse set of evaluation metrics, namely the $\Fone$ score, the Matthews correlation coefficient, Cohen’s kappa, and accuracy.
In Section~\ref{GANs-Results}, based on a framework incorporating long short‑term memory and generative adversarial structure, knowledge from human-experience-trained algorithms are incorporated to generate textual data. 
Section~\ref{Conclued} concludes the paper.

\section{Data}
\label{data}
\begin{figure}[h!]
\centering
{\includegraphics[width=0.95\textwidth]{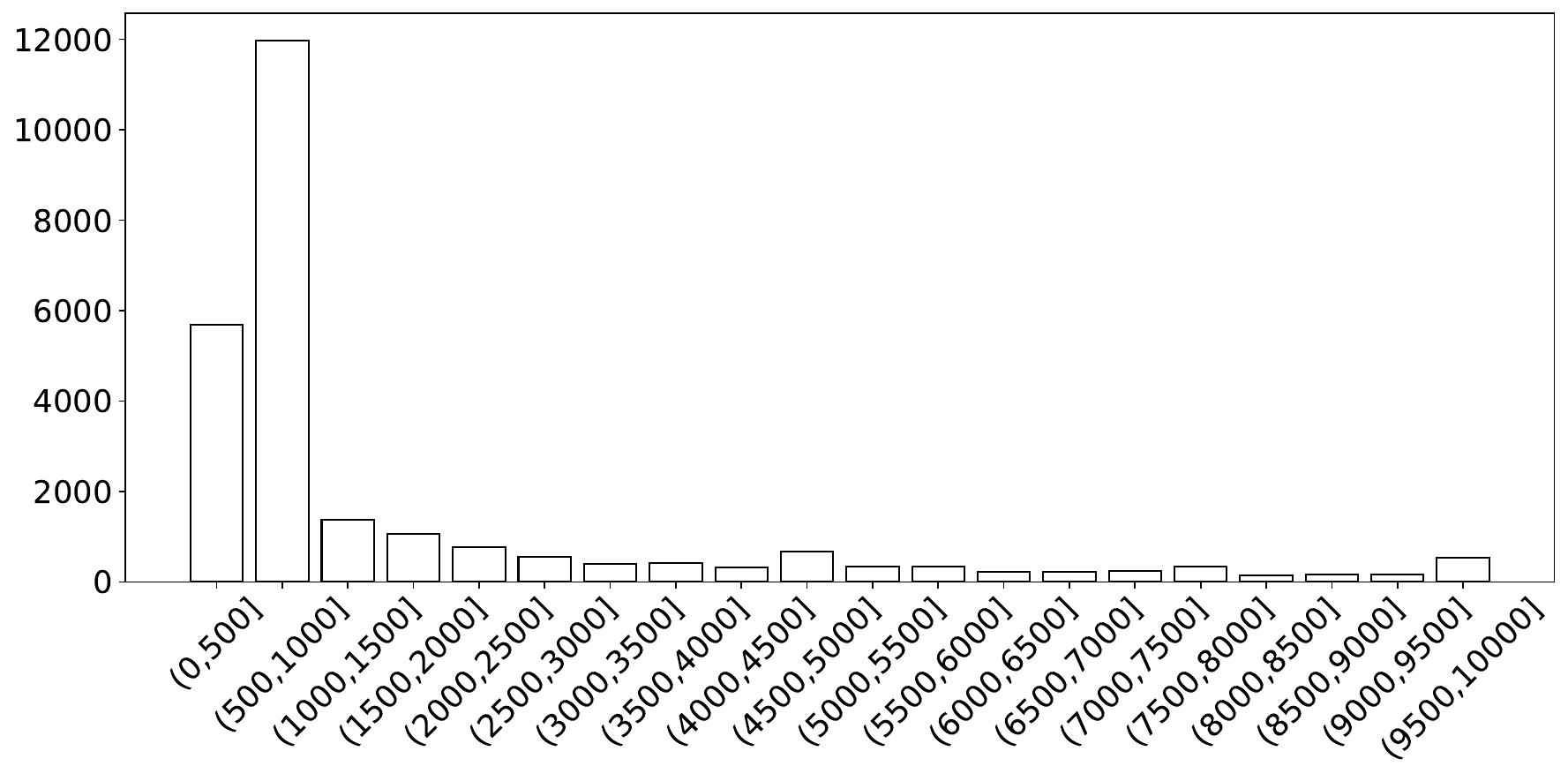}}

\caption{Frequencies of the complaint narratives in various dollar value.}
\label{Dollar_valuePlot}
\end{figure}

We utilize a dataset similar to that of \citet{Gao2025}. However, by involving more claims in our analysis, we tend to work with a larger dataset. We select only those complaints that include the narratives satisfying the following three properties:
\begin{enumerate}
    \item The claims were submitted in last 5-year between January 1, 2020, and December 31, 2024.
    \item The claims labeled as ``Closed with explanation", ``Closed with non-monetary relief", and ``Closed with monetary relief" are selected. Those claims have been resolved with monetary or non-monetary reliefs are labeled as ``meritorious", while those resolved solely through responses are classified as ``non-meritorious" \citep{HJP2021}.
    \item The claims are selected with a frequency higher than 1,000 under the categorical columns (e.g., product, issue, company).
\end{enumerate}
In the selected complaint records, only those claims that refer to dollar amounts larger than \$0 are considered, and we remove the claims with dollar amounts exceeding \$10,000, because such claims are more likely to be associated with business accounts rather than individual accounts \citet{Gao2024}. Hence, these steps result in the selection of 31,036 complaints for further analysis. Among these, 15,052 are deemed meritorious, and the remaining 15,984 complaints are deemed non-meritorious. Since the number of data samples labeled as meritorious and non-meritorious is nearly equal, oversampling techniques are not considered in the following analysis.
For each distinct narrative has been required to include the following details: ``Data received", ``Product", ``Issue", ``Company", ``Consumer complaint narrative", ``Dollar value", and ``Meritorious". Among these seven features in the corpus, the ``Dollar value" is a numerical column that captures the dollar amounts extracted from the ``Consumer complaint narrative" textual column. 
Figure~\ref{Dollar_valuePlot} 
depicts the ``Dollar value" column from smallest to largest and divided into twenty groups, each corresponding to an equal range of dollar amounts. Notably, over 90\% of complaints involve smaller dollar amounts, specifically within the \((\$0, \$1,000]\) range. This trend indicates that consumer grievances are predominantly associated with modest financial claims. In Figure~\ref{Dollar_valuePlot}, the distribution of the dollar value is heavily right-tailed; consequently, we apply a logarithmic transformation to this numerical column. This transformation helps stabilize the variance and reduce skewness, making the data more normally distributed and improving the reliability of subsequent statistical analyses.

In the dataset, the columns labeled ``Product," ``Issue," and ``Company" serve as the categorical features, while the column labeled ``Meritorious" functions as the target categorical variable. The column labeled ``Consumer complaint narrative" is a textual column that contains valuable information for our subsequent analysis. To incorporate this data effectively, text cleaning is crucial; it is a critical pre-processing step that enhances the quality and consistency of the data, thereby improving the reliability and accuracy of subsequent analyses. Hence, the following steps have been implemented to clean the narrative within the corpus:

\begin{enumerate}
    \item Removing redundant whitespace, i.e., merging multiple consecutive spaces or lines into one if they do not serve a purpose.
    \item Removing empty documents or texts that contain almost no useful content, i.e., extremely short texts with only one, two words in complaints.
    \item Removing special characters that cannot be parsed or have no semantic meaning. (i.e., XXXX, XXXX/XX/XX)
\end{enumerate}

For better analysis of the textual dataset, normally the corpus with shorter length yield the better performance than the longer length \citep{KK2022,Siddharthan2014}. For complaint data, text lengths can vary. To make each complaint comparable, we use the T5-base pre-trained model provided by Google \citep{RSRLNMZL2020} to summarize the text to a specified length. 
We restrict the length of each summarized text to a maximum of 128 words (150 tokens) in order to facilitate a more rigorous and detailed analysis as recommend by \citet{RA2022}. 

To assess how closely our generated summaries match the source texts in meaning, we measure the cosine similarity between their vectorized representations. Following \citet{vgpz2023}, we first convert the original document and its T5‑base summary into bag‑of‑words vectors based on term frequencies. Cosine similarity then evaluates the cosine of the angle between these high‑dimensional vectors, yielding a value between zero (no overlap) and one (perfect alignment). Because it depends only on the orientation of vectors, this scale‑invariant metric emphasizes semantic correspondence independently of text length or absolute word counts.

Suppose we randomly select a narrative ($\mathbf{s}_o$) from our corpus and use the pre-trained T5-base model to summarize its corresponding textual sentence ($\mathbf{s}_s$) as:
\begin{equation*}
\begin{aligned}
\mathbf{s}_o &= \textit{unexpected fees charged to my credit account and inaccurate statements}, \\
\mathbf{s}_s &= \textit{unexpected fees and inaccurate credit statements}.
\end{aligned}
\end{equation*}
Table~\ref{tab:vectors} 
\begin{table}[h]
\centering
\caption{Word frequencies between $\mathbf{s}_o$ and $\mathbf{s}_s$.}
\begin{tabular}{c|cccccccccc}
 & account & and & charge & credit & fees & inaccurate & my & statements & to & unwanted \\
\hline
$\mathbf{s}_o$ & 1 & 1 & 1 & 1 & 1 & 1 & 1 & 1 & 1 & 1 \\
$\mathbf{s}_s$ & 0 & 1 & 0 & 1 & 1 & 1 & 0 & 1 & 0 & 1 \\
\end{tabular}
\label{tab:vectors}
\end{table}
presents the first step in computing cosine similarity: collecting all unique words in \(\mathbf{s}_o\) and \(\mathbf{s}_s\), and then counting the frequency of each word. Thus, the cosine similarity ($\mathcal{CS}$), defined as the cosine of the angle between the two vectors, is determined to be
\[
\mathcal{CS}(\mathbf{s}_o, \mathbf{s}_s) = \frac{\mathbf{s}_o \cdot \mathbf{s}_s}{\|\mathbf{s}_o\| \|\mathbf{s}_s\|} = \frac{(1, 1, 1, 1, 1, 1, 1, 1, 1, 1)\cdot (0, 1, 0, 1, 1, 1, 0, 1, 0, 1)}{\sqrt{10}\sqrt{6}} \approx 0.775.
\]
The calculated cosine similarity, approximately 0.775, indicates a high degree of semantic alignment between the summary and the original sentence. This result implies that the summary effectively captures the essential content of the original while omitting less critical details. 

\begin{figure}[h!]
\centering
\subfigure[\text{Raw narratives.}]{\includegraphics[width=\textwidth]{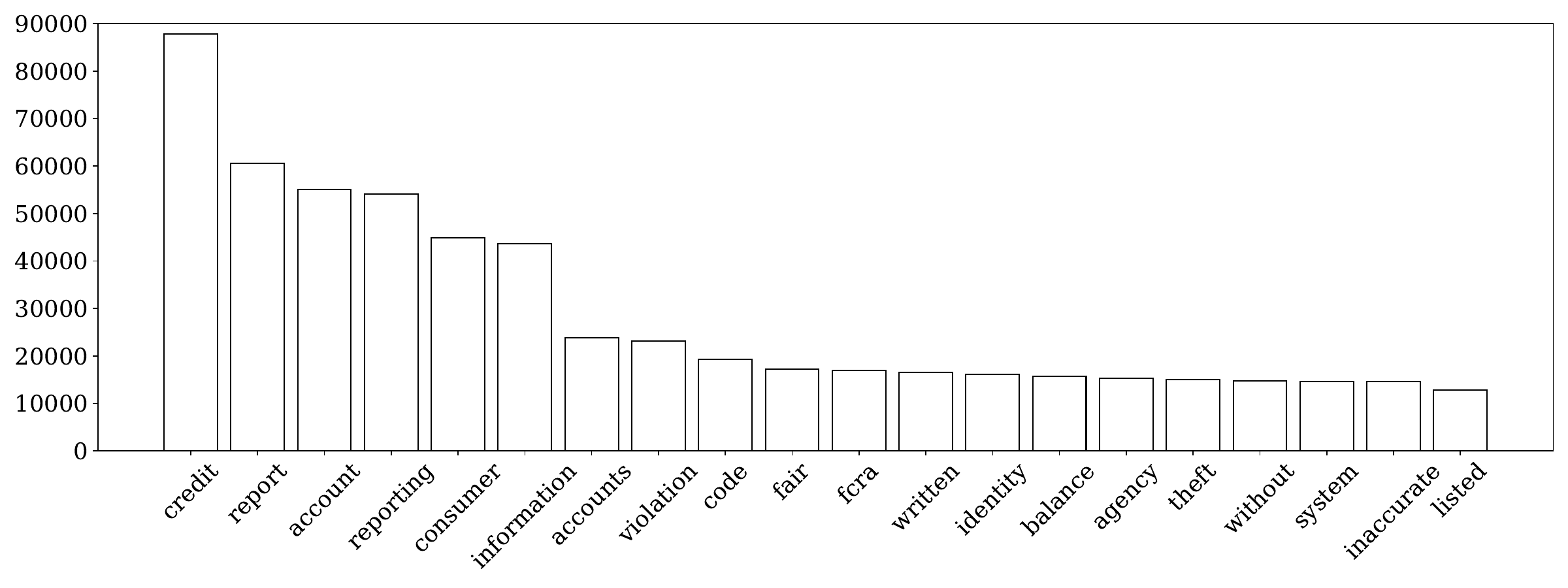}}
\subfigure[\text{Summarized text with maximum 128 words.}]{\includegraphics[width=\textwidth]{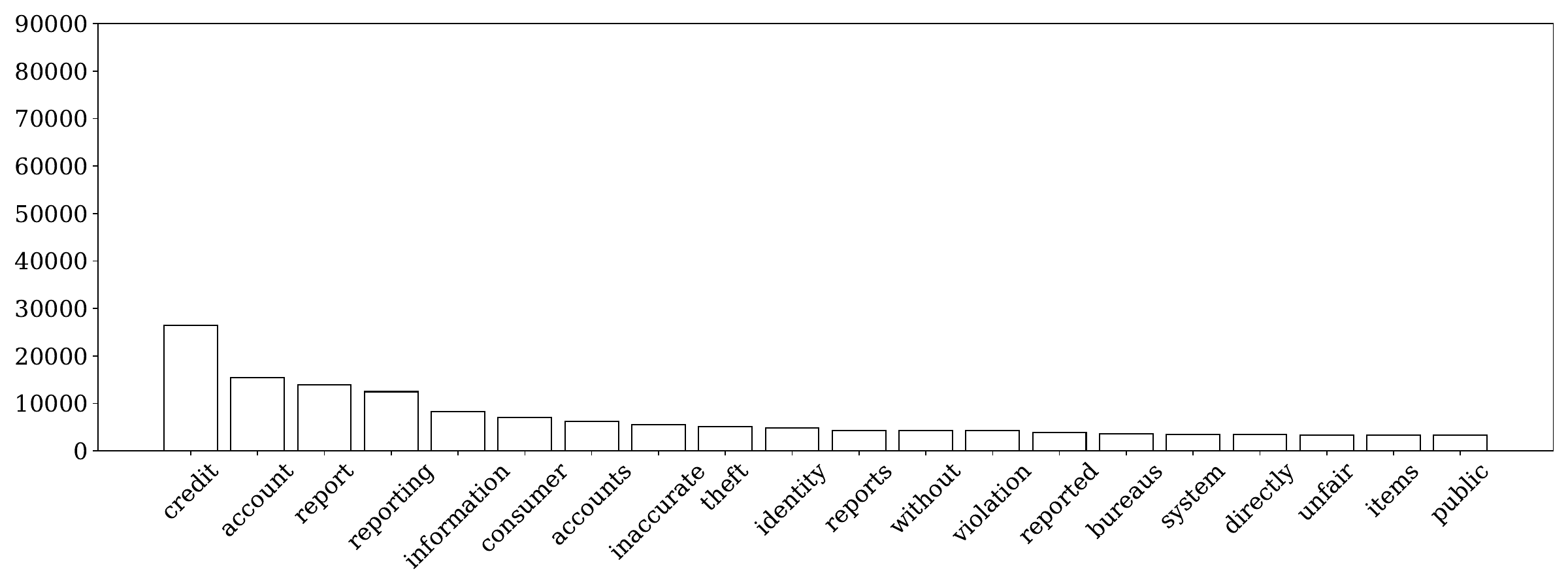}}

\caption{Comparison of keywords before (top) and after (bottom) summarization.}
\label{WordClould_plot}
\end{figure}
Figure~\ref{WordClould_plot} compares the histograms of word importance before and after applying text summarization (e.g., T5-based model), revealing that although the summarization process filters out less relevant words, the core keywords and underlying sentiments remain largely unchanged. This observation underscores the capability of advanced T5-based model to distill extensive texts into their most informative components, preserving the semantic integrity and emotional cues of the original content. Such distilled representations reduce noise and redundancy and enhance the overall quality of the features used in subsequent analytical tasks. Moreover, the resulting concise text representations facilitate the conversion of text into numerical matrices for classification, thereby improving computational efficiency while ensuring that critical insights are effectively retained.

\section{Converting narratives into quantitative data}
\label{Narrative_classfication}

In the current section, we incorporate all features into our analysis (e.g., the textual component and the categorical components). The textual column is arguably the most critical element in our dataset, as it contains valuable semantic information. We first transform the textual data into a numerical representation in the following analysis. This representation is then integrate with the categorical components and use as input for our machine learning models. We evaluate the performance of different ML classifiers using a range of evaluation metrics.

\subsection{Text featurization approaches}
Text featurization, the initial step in many classification tasks, is a fundamental process that transforms unstructured textual data into a structured representation, thereby facilitating efficient processing by ML algorithms. Extracting meaningful features uncovers the intrinsic patterns and contextual relationships within the text. We next implement the following two novel text featurization approaches of textual data into numerical matrices.

\subsubsection{Truncated term frequency–inverse document frequency (Truncated TF-IDF)}
TF-IDF is a statistical technique that evaluates the importance of individual words within a document relative to their frequency across the entire corpus. Variants of TF-IDF are widely utilized in information retrieval applications, such as document classification, scoring, and ranking the relevance of documents in response to search queries.

Formally, using the notation of \citet{Gao2024}, let $\mathcal{C}=\{c_1,\ldots,c_n\}$ be the set of all cleaned complaint narratives and $\mathcal{W}$ the set of all words appearing in $\mathcal{C}$. Then the term frequency $\text{TF}:\mathcal{C}\times\mathcal{W}\mapsto\mathbb{N}$ is \citep{s1972} the frequency of the $j^{\text{th}}$ word $w_j$ in the $i^{\text{th}}$ cleaned complaint narrative $c_i$. The inverse document frequency   $\text{IDF}: \mathcal{W}\mapsto[0,\infty)$ is defined as
$$\text{IDF}(w_j) = \log\bigg(\frac{n}{n_j}\bigg)$$
for $w_j\in\mathcal{W}$, where 
$n_j = \sum^n_{i=1}\mathds{1}\{\text{TF}(c_i, w_j) > 0\}$ 
is the number of cleaned narratives that include $w_j$. Then $\text{TF-IDF}: \mathcal{C}\times\mathcal{W}\mapsto[0,\infty)$ is given by \citep{blg1998}  
$$\text{TF-IDF}(c_i, w_j) = \text{TF}(c_i, w_j)\times \text{IDF}(w_j).$$
Using the TF-IDF word embedding process, the cleaned complaint narratives are translated into a matrix with $n$ rows and $\lvert \mathcal{W}\rvert$ columns. 

When applying TF-IDF to a huge corpus (e.g., $31,036$ number of narratives), each document is represented in a high-dimensional space where the dimensions correspond to every unique word in the entire collection. Since any individual document contains only a small subset of these words, most entries in the resulting document-term matrix are zero. This high level of sparsity is inherent to the nature of textual data, where vocabulary size grows significantly with a large corpus. Yet, each document only uses a fraction of the total vocabulary.

To address the challenges associated with sparse matrices, the matrix TF-IDF has the singular value decomposition (SVD) form. Following the recommendations of \citet{HWY2023}, we reduced the dimensionality to a 768-dimensional vector. Given a TF-IDF matrix
\[
\text{TF-IDF}(c_i, w_j) \in \mathbb{R}^{n \times \lvert \mathcal{W}\rvert},
\]
by applying the knowledge about SVD. This factorizes \(\text{TF-IDF}(c_i, w_j)\) as:
\[
\text{TF-IDF}(c_i, w_j) \approx U \Sigma V^T,
\]
where \(U \in \mathbb{R}^{n \times 768}\) is the matrix of left singular vectors,  \(\Sigma \in \mathbb{R}^{768 \times 768}\) is a diagonal matrix containing the top \(768\) singular values, and \(V \in \mathbb{R}^{\lvert \mathcal{W}\rvert \times 768}\) is the matrix of right singular vectors.

The new TF-IDF matrix with smaller dimensions is known as truncated TF-IDF, which seeks to reduce the disproportionate influence of common terms. Under this featurization approach, the term frequency component is limited to a predetermined threshold, thus decreasing the weight assigned to certain words. The resulting metric offers a more balanced representation of text data, improving task performance.

\subsubsection{$\text{FinBERT-IDF}_{\text{token}}$}
TF-IDF is a widely used method for feature representation at the word level, often producing strong results for short texts with relatively small vocabularies. Its effectiveness mainly arises from its ability to measure the importance of a term in a document compared to its frequency in the entire corpus. However, as the corpus size increases, the vocabulary expands significantly, leading to higher dimensionality in the feature space, and consequently, increasingly sparse TF-IDF matrices. This sparsity not only complicates subsequent analyses but also presents challenges for many machine learning algorithms that find it difficult to learn effectively from high-dimensional, sparse data.

Furthermore, the vocabulary size can expand dramatically when transitioning from word-level to subword-level feature representation (e.g., by breaking down each word into smaller tokens). This expansion further worsens the sparsity issue, particularly in cases where languages with complex morphological structures or subword segmentation techniques are utilized. Although dimensionality reduction techniques such as singular value decomposition can be used to compress the high-dimensional TF-IDF matrix, these methods are essentially linear approximations and do not address the fundamental limitation of TF-IDF as a static statistical measure.

Hence, a deep featurezation approach (e.g., FinBERT) could be necessary. FinBERT \citep{HWY2023} is a transformer model pre‑trained on general text and then fine‑tuned on financial documents to yield rich, context‑sensitive embeddings.  Given an input sequence of \(n\) tokens \(\mathcal{S}=\{x_1,\dots,x_n\}\), each  initial embedding of token is formed by summing three components:  
\[
\mathbf{x}_i \;=\;\mathbf{e}_{\text{token}}(x_i)\;+\;\mathbf{e}_{\text{pos}}(i)\;+\;\mathbf{e}_{\text{seg}}(x_i).
\]  
These embeddings pass through 12 transformer layers to produce an output matrix \(H\), from which the \(\texttt{[CLS]}\) row serves as the 768‑dimensional sentence vector:  
\[
\mathbf{v}\;=\;H^{(12)}_{\text{[CLS]}}\;\in\;\mathbb{R}^{768}.
\]  
Applied to our dataset of 31,036 consumer complaints as discussed in Section~\ref{data}, this yields the feature matrix  
\(\mathbf{V} \in \mathbb{R}^{31{,}036\times768}\).  

Although these embeddings capture deep contextual information, they can still suffer from high dimensionality and sparsity.  To address this, we propose $\text{FinBERT-IDF}_{\text{token}}$, which reweights each token embedding by its inverse document frequency (IDF), thereby amplifying the contribution of rare but informative terms.  Concretely, we compute  
\[
\tilde{\mathbf{e}}_{\text{token}}(x_i)
=\mathbf{e}_{\text{token}}(x_i)\times\mathrm{IDF}(x_i),
\]
and then redefine the token embedding as  
\[
\mathbf{x}_i
=\tilde{\mathbf{e}}_{\text{token}}(x_i)
+\mathbf{e}_{\text{pos}}(i)
+\mathbf{e}_{\text{seg}}(x_i).
\]  
By incorporating corpus‑level weighting into the transformer pipeline, $\text{FinBERT-IDF}_{\text{token}}$ reduces sparsity and preserves nuanced, domain‑specific patterns, making it particularly effective for complex financial text classification.

\subsection{Machine learning classifiers}
\label{Real_data_Classfication}
Traditional models (e.g., logistic regression (LR) and random forest (RF)) often perform well when data are simple and structured, delivering accurate classification with minimal complexity. However, as the variability and complexity of the data increase, these basic models may struggle to capture subtle nuances and contextual dependencies. In such cases, deep learning models like LSTM and FinBERT become essential. They are designed to learn sophisticated language representations that automatically discern intricate semantic patterns. This capability enhances classification accuracy and offers greater robustness in handling diverse and complex text data.

In this study, we utilize both traditional models and deep learning architectures. Traditional models, recognized for their computational efficiency and interpretability, perform adequately on straightforward, well-structured datasets; however, they often struggle to capture the nuanced semantic and syntactic relationships present in more complex textual data. In contrast, deep learning architectures like FinBERT excel at extracting intricate language patterns from unstructured inputs. Additionally, we incorporate an LSTM model, which demonstrated excellent performance in \citet{Gao2025}.  Furthermore, the advanced FinBERT model is implemented on a high-performance GPU, with an M4 MAX GPU used to expedite data preprocessing. The objective of this research is to identify the classifier that exhibits optimal performance in analyzing complex textual data, such as consumer complaints.

\subsection{Evaluation metrics}
\label{Eva_met}

Using the truncated TF‑IDF and $\text{FinBERT-IDF}_{\text{token}}$ featurization methods, textual data can be represented as a 768‑dimensional numerical matrix. After selecting appropriate machine‑learning classifiers, the crucial final step is to choose a suitable evaluation metric (standard), which is essential for accurately assessing the performance of models.

In \citet{Gao2024}, we focus on accuracy as our primary evaluation standard. This standard (accuracy) is simple and interpretable, particularly when applied to balanced datasets with small size. 
In domains like consumer complaint analysis, increasing the number of complaints and including additional subgroups often leads to a significant class imbalance (for example, relatively few narratives are labeled meritorious). In \citet{Gao2025}, we shift our focus to the $\Fone$ score as the main evaluation standard. Using the $\Fone$ score provides a more balanced performance measure in the presence of class imbalance.

In this study, building on our understanding of accuracy and the $\Fone$ score, we introduce two additional evaluation standards to deepen our insight into the classification task. The response variable is binary, distinguishing between meritorious and non‑meritorious cases. \citet{CWJ2021} demonstrate that the $\MCC$ delivers the most robust assessment in binary classification by accounting for true positives, true negatives, false positives, and false negatives. The $\MCC$ is computed as
\begin{equation}
    \MCC = \frac{(\TP \times \TN) - (\FP \times \FN)}{\sqrt{(\TP + \FP)(\TP + \FN)(\TN + \FP)(\TN + \FN)}}, \qquad \MCC\in[-1,1]
\end{equation}
when $\MCC = 1$, it indicates perfect prediction, 0 indicates random prediction, and -1 indicates complete disagreement between predictions and actual values.

As our input variables consist of a combination of categorical and textual variables, \citet{KPKWH2024} and \citet{KK2022} conclude that Cohen's kappa (\(\mathrm{C}_{\kappa}\)), which measures the level of agreement between two raters while accounting for the possibility of chance agreement, demonstrates the best performance in NLP. The \(\mathrm{C}_{\kappa}\) is computed as
\begin{equation}
    \mathrm{C}_{\kappa} = \frac{p_o - p_e}{1 - p_e},\qquad \mathrm{C}_{\kappa}\in[-1,1]
\end{equation}
where \(p_o\) is the observed agreement (accuracy), and \(p_e\) is the expected agreement by chance (expected accuracy). 

In the subsequent analysis, we employ accuracy, the $\mathrm{F}_1$ score, the Matthews correlation coefficient, and Cohen’s kappa as our evaluation metrics. Accordingly, we define the metric set as  
$\mathcal{E} = \{\Acc,\;\Fone,\;\MCC,\;\mathrm{C}_{\kappa}\}.$  Algorithm~\ref{algorithm-F1_computed}
\begin{table}[h!]  
\centering
\caption{Performance of diverse evaluation metrics in the NLP-based assessment of textual consumer complaints.}
\begin{tabular}{l}
\hline 
{\bf Require} 
\\
{1.} A dataset $D$ with true labels ``meritorious" and ``non-meritorious". \\
{2.} A specific classfication method. \\
\hline 
{\bf Compute the evaluation metrics $\mathcal{E}$ }  
\\
{1.} Convert the input categorical variables to a dummy variables named $X_{\text{dummy}}$.
\\
{2.} Convert the input textual variables to a numerical matrix based on the text featurization 
\\ \quad approach, and append the numerical matrix to $X_{\text{dummy}}$.
\\
{3.} Randomly assign (1-r)100\% of the dataset $D$ as a training set, and the remaining r100\% 
\\ \quad as a testing set.
\\
{4.} Train an ML classifier $f$ based on the training set.
\\
{5.} Apply $f$ on the testing set and compute the ML label.
\\
{6.} Compute the evaluation metrics.

\\ \hline 
\end{tabular}
\renewcommand{\tablename}{Algorithm} 
\label{algorithm-F1_computed}
\end{table}
 outlines the procedure by which we apply the input variables to the machine‑learning classifiers and compute various evaluation metrics. In the following section, we illustrate classifier performance using Algorithm~\ref{algorithm-F1_computed} and employ both TF‑IDF (as discussed in \citet{Gao2024}) and FinBERT featurization (as suggested by \citet{HWY2023}) to benchmark our proposed methods.

\subsection{Classification results in real narratives}
\label{Classfication_result}
In traditional word embedding techniques like TF‑IDF, each word is treated as an independent term without considering context or word order. This method counts the frequency of words and scales them by their inverse document frequency, meaning all words are assumed to contribute equally, regardless of their position or relationship with other words. In contrast, deep learning models such as LSTM process text as a sequence, capturing the contextual dependencies and subtle semantic relationships between words. This fundamental difference in how information is represented can lead to inconsistencies when combining the two approaches. For instance, feeding TF‑IDF vectors into an LSTM does not make sense because the sequential information that the LSTM is designed to exploit is absent in a bag-of-words representation. As a result, the performance metrics derived from such a hybrid approach yield results that are not incorporated into the research.

\begin{table}[h!]
\centering
\caption{Performance summary in the percentage of ML classifiers on consumer narratives.}
\begin{tabular}{clrrrr}
\hline\hline 
\multirow{2}{*}{Evaluation metrics} & \multirow{2}{*}{Featurization approaches} & \multicolumn{4}{c}{ML classifiers} 
\\ 
\cline{3-6}
& & LR~ & RF~ & LSTM  & FinBERT \\ \hline
\multirow{4}{*}{$\Acc$} & TF-IDF & 59.78 & 58.20 & -  & - \\ 
& Truncated TF-IDF & 59.54 & 56.38 & -  & - \\ 
& FinBERT & 65.80 & 72.31 & 77.70  & 77.45 \\ 
& $\text{FinBERT-IDF}_{\text{token}}$ & 65.00 & 71.81 & 77.97  & 77.63 \\ \hline
\multirow{4}{*}{$\Fone$ score} & TF-IDF & 59.77 & 58.21 & -  & - \\ 
& Truncated TF-IDF & 59.53 & 56.36 & -  & - \\ 
& FinBERT & 65.81 & 72.25 & 77.69  & 77.41 \\ 
& $\text{FinBERT-IDF}_{\text{token}}$ & 63.51 & 69.88 & 77.72  & 77.58  \\\hline
\multirow{4}{*}{$\MCC$} & TF-IDF & 19.63 & 16.43 & -  & - \\ 
& Truncated TF-IDF & 19.08 & 12.69 & -  & - \\ 
& FinBERT & 31.55 & 44.56 & 55.36  & 54.86 \\ 
& $\text{FinBERT-IDF}_{\text{token}}$ & 29.89 & 43.56 & 55.98  & 55.37 \\\hline
\multirow{4}{*}{$\mathrm{C}_{\kappa}$} & TF-IDF & 19.60 & 16.42 & -  & - \\ 
& Truncated TF-IDF & 19.07 & 12.68 & -  & - \\ 
& FinBERT & 31.55 & 44.46 & 55.33  & 54.78 \\ 
& $\text{FinBERT-IDF}_{\text{token}}$ & 29.89 & 43.45 & 55.94  & 55.31  \\\hline
\end{tabular}
\label{table1-true}
\end{table}
In Table~\ref{table1-true}, this incompatibility is acknowledged by marking the relevant entries with a dash ``–” in our results table, indicating that although performance was measured, the results do not accurately reflect the underlying capability of models. We present a comparative evaluation of different ML classifiers applied to consumer narratives using various featurization approaches in Table~\ref{table1-true}.  Traditional methods such as TF‑IDF and truncated TF‑IDF, which treat each word as an independent term, are only compatible with simpler classifiers like LR and RF. Empirical evidence indicates that these methods exhibit moderate performance, achieving accuracy figures in the upper 50\% range and corresponding F1 scores, while producing relatively low values for the $\MCC$ and $\mathrm{C}_{\kappa}$ metrics.

In contrast, all four evaluation metrics exhibit a similar pattern: as the accuracy of the model increases, comparable trends are observed for the other three evaluation metrics (\(\Fone\), \(\MCC\), and \(\mathrm{C}_{\kappa}\)). Furthermore, the truncated TF-IDF featurization approach yields performance comparable to the traditional TF-IDF method, which is derived from TF-IDF by implementing singular value decomposition.

When employing both FinBERT and \(\text{FinBERT-IDF}_{\text{token}}\) for featurization, significant variability is observed across classifiers. For simpler models, such as logistic regression (LR) and random forest (RF) classifiers, the FinBERT featurization approach achieves the highest performance, with accuracies of 65.80\% for LR classifiers and 72.31\% for RF classifiers. In contrast, for deep learning models, such as LSTM and FinBERT classifiers, the \(\text{FinBERT-IDF}_{\text{token}}\) featurization approach attains comparable performance, yielding accuracies of 77.97\% for LSTM classifiers and 77.63\% for FinBERT classifiers. These findings suggest that the \(\text{FinBERT-IDF}_{\text{token}}\) approach provides robust performance when deep learning models are employed, as it effectively integrates the advantages of both FinBERT and TF-IDF methodologies. However, a notable drawback of this novel featurization approach is its increased computational time relative to FinBERT, particularly when analyzing large datasets, due to its more comprehensive textual information aggregation.

The results show that contextualized embeddings, especially those tailored for the financial domain, such as FinBERT and \(\text{FinBERT-IDF}_{\text{token}}\), are more effective in capturing the nuances of consumer narratives than traditional TF-IDF representations. In our subsequent analysis, we focus on implementing the FinBERT featurization approach due to its superior time efficiency.

\section{Text generation in human-experience-trained algorithms}
\label{GANs-Results}
In Section~\ref{Real_data_Classfication}, we have demonstrated that integrating deep learning models into textual analysis yields significantly enhanced classification results. Although our initial experiments are conducted on an existing dataset, the use of real-world data in practice is often constrained by high acquisition costs, data privacy issues, and inconsistencies in data quality. To address these limitations, we have opted to generate a text dataset. This approach circumvents the practical challenges associated with real data and allows for greater control over experimental conditions, thereby facilitating systematic evaluations and reproducibility of our results.

\subsection{Text generation}
Text generation is a subfield of NLP that involves creating coherent and contextually relevant sequences of text based on learned patterns from training data. The process typically begins with converting raw textual data into a numerical representation, which enables machine learning models to process and understand the data.  Not all numerical representation methods are suitable for generation. For instance, techniques like TF-IDF are primarily designed for feature extraction. They compute term importance based on frequency and inverse document frequency without accounting for word order or contextual dependencies. TF-IDF operates under a bag-of-words assumption that disregards the sequential structure of language, rendering it incapable of capturing the syntactic and temporal relationships necessary for generating coherent text.

Two principal approaches to converting raw textual data into a numerical representation for text generation have traditionally been pursued: statistical n-gram model and methods based on text embeddings. An N-gram model operates by computing the conditional probability of a word given a fixed window of preceding words, thereby offering a straightforward and interpretable means to generate text based solely on local context frequencies.  \citet{Garg2022} explains the importance of bigram model for feature selection from the short text.

A bigram (e.g., n =2) model is a statistical language model that predicts the next word in a sequence based on its immediate predecessor. It achieves this by analyzing the frequency of word pairs in a given corpus and computing the corresponding conditional probabilities. For instance, we select one of the summarized sentences from the \citet{CFPB}:
\begin{center}
    \textit{unexpected fees and inaccurate credit statements}
\end{center}
when processed using a bigram approach, it can be partitioned into the following pairs: (``unexpected",``fees"), (``fees", ``and"), and so on.  
Table~\ref{Bigram} 
\begin{table}[h]
    \centering
    \caption{Bigram frequencies and probabilities for one sentence from the CFPB.}
    \begin{tabular}{lcc}
        Bigram & Frequency &  \(P(w_i \mid w_{i-1})\) \\
        \hline \hline
        (``unexpected", ``fees") & 1 & 1.0 \\
        (``fees", ``and") & 1 & 1.0 \\
        (``and", ``inaccurate") & 1 & 1.0 \\
        (``inaccurate", ``credit") & 1 & 1.0 \\
        (``credit", ``statements") & 1 & 1.0 \\
        \hline
    \end{tabular}
    \label{Bigram}
\end{table}
presents the frequencies and probabilities derived from the example sentence. Each row in the table represents a unique bigram (a pair of consecutive words) along with its frequency of occurrence and the corresponding conditional probability \(P(w_i \mid w_{i-1})\). For instance, the bigram (``unexpected", ``fees") occurs once in the corpus, and the probability of ``fees" following ``unexpected" is calculated as \(1.0\). This table serves as the foundation for a bigram language model, enabling the prediction of the next word in a sequence based on the previous word. By analyzing such probabilities, the model can generate coherent text or estimate the likelihood of word sequences in natural language processing tasks.

The second approach is text embeddings, which transform discrete textual elements into continuous vector representations that capture semantic and syntactic nuances. Three primary sub-areas are widely recognized under this umbrella: character embeddings, word embeddings, and subword tokenization. Character embeddings represent individual characters as vectors, providing fine-grained control and robustness to morphological variations and out-of-vocabulary terms. Word embeddings, on the other hand, assign a unique dense vector to each word, effectively encapsulating overall semantic and syntactic properties. 

Subword tokenization is a technique that decomposes words into smaller, semantically meaningful units. This method enables language models to handle rare or unseen words efficiently and reduces input sequence lengths, enhancing computational performance and preserving essential linguistic nuances. This approach underpins many state-of-the-art pre-trained models, such as FinBERT and T5, which employ subword tokenization as a core architecture component. These models can be fine-tuned for specific tasks or domains, enabling rapid adaptation and improved performance in applications ranging from financial sentiment analysis to text generation. 

After converting raw textual data into a numerical representation, the subsequent step involves selecting an LSTM architecture for text generation. LSTM networks constitute a recurrent neural network specifically designed to process sequential data. They employ memory cells that preserve crucial information over time and incorporate gating mechanisms to regulate which details are retained or discarded.

By providing numerical representations as inputs to both a text generation model that utilizes long short-term memory and a statistical bigram model, we can generate brief complaints (i.e., those containing 10 words) in which every sentence commences with the word ``credit”. Table~\ref{text-embedding-LSTM} 
\begin{table}[h]
\centering
\caption{Narratives generation via statistical bigram and subword tokenization models with meritorious indicators.}
\begin{tabular}{llc}
\hline \hline
Model  & Narratives generation & Meritorious \\
\hline \hline
Statistical bigram   & \begin{tabular}[c]{@{}l@{}}credit report fraudulent accounts obtain \\ credit card account remains unresolved \end{tabular} & Yes \\
\hline
Statistical bigram   & \begin{tabular}[c]{@{}l@{}}credit file dont belong would \\ removed credit card account ending \end{tabular} & No\\
\hline
Subword tokenization-FinBERT  & \begin{tabular}[c]{@{}l@{}}credit legitimacy complies infringed  primary \\ amounts thought carolina represents inquire\end{tabular}& Yes \\
\hline
Subword tokenization-FinBERT &  \begin{tabular}[c]{@{}l@{}}credit bureaus tax audit acquiring  aims \\ correspondence neglecting location changed\end{tabular} & No \\
\hline
\end{tabular}
\label{text-embedding-LSTM}
\end{table}
demonstrates that the subword tokenization-FinBERT model produces more consistent outputs than the statistical bigram model. Subword segmentation of FinBERT effectively handles out-of-vocabulary issues and captures complex compound expressions in financial language. In contrast, the model of bigram reliance on fixed word pair frequencies can lead to occasional inconsistencies and a weaker contextual understanding.

\subsection{Generative adversarial-based human-experience-trained algorithms}
LSTM networks are extensively used for text generation because they effectively capture long-term dependencies in sequential data. However, a discrepancy arises between training and inference: during training, the model uses the true previous word as context, whereas during inference, it relies on its own generated words. This difference can lead to error accumulation, as the model may encounter unfamiliar contexts if it generates an unlikely word, potentially resulting in incoherent outputs. Essentially, the model becomes overly dependent on the ideal contexts present during training and is less resilient to its own mistakes. Adversarial training with GAN \citep{GPMBWD2014} addresses this issue by having the LSTM generator produce entire sequences in a free-running mode (using its outputs recursively, as in inference).

The GAN is the gaming (with respect to the expectation of the log utility) between the generator and discriminator, namely,
$$\min_{\boldsymbol{\Theta}, \boldsymbol{\psi}}\max_{\boldsymbol{\phi}\in\mathbb{R}^{769}} V({\rm LSTM}_{\boldsymbol{\Theta}, \boldsymbol{\psi}}, D_{\boldsymbol{\phi}})$$
where $D_{\boldsymbol{\phi}}:\mathbb{R}^{768}\mapsto(0, 1)$ is defined as the igmoid function on the inner product of a 769-dimensional vector $\boldsymbol{\phi}$ and another 768+1-dimensional vector $\begin{bmatrix} \boldsymbol{h}^{\intercal} & 1\end{bmatrix}^{\intercal}$:
$$D_{\boldsymbol{\phi}}(\boldsymbol{h}) := \sigma\left(\begin{bmatrix} \boldsymbol{h}^{\intercal} & 1\end{bmatrix}\boldsymbol{\phi}\right) \triangleq \frac{1}{1 + \exp\left(-\begin{bmatrix} \boldsymbol{h}^{\intercal} & 1\end{bmatrix}\boldsymbol{\phi}\right)}$$
and
$$V({\rm LSTM}_{\boldsymbol{\Theta}, \boldsymbol{\psi}}, D_{\boldsymbol{\phi}}) = \mathbb{E}_{\boldsymbol{x}}[\log D_{\boldsymbol{\phi}}(\boldsymbol{h}^{{\rm LSTM}_{\boldsymbol{\Theta}, \boldsymbol{\psi}}})] + \mathbb{E}_{\boldsymbol{z}}[\log(1 - D_{\boldsymbol{\phi}}(\boldsymbol{h}^{{\rm LSTM}_{\boldsymbol{\Theta}, \boldsymbol{\psi}}}))].$$
The hidden layer output $\boldsymbol{h}$ of the observed narrative $\boldsymbol{x} = \{x_t\}^{T_x}_{t=1}$ is determined recursively through
$$\boldsymbol{h}_t = {\rm LSTM}_{\boldsymbol{\Theta}, \boldsymbol{\psi}}(x_t, \boldsymbol{h}_{t - 1})$$
for $t = 1,2,\ldots$ and $\boldsymbol{h}_0 = \boldsymbol{0}$ while the hidden layer output $\boldsymbol{h}$ of the generated narrative $\boldsymbol{z} = \{z_t\}^{\infty}_{t=1}$ is determined through
$$\boldsymbol{h}_t = {\rm LSTM}_{\boldsymbol{\Theta}, \boldsymbol{\psi}}(z_t, \boldsymbol{h}_{t - 1})$$
with
$$p(z_t|\boldsymbol{h}_{t-1}) = {\rm Softmax}\left(\begin{bmatrix} \boldsymbol{h}^{\intercal}_{t - 1} & 1\end{bmatrix}\boldsymbol{\psi}\right)$$
for $t = 1,\ldots,\tau$ where $\tau$ is some stopping time.

Subsequently, the next step involves incorporating an LSTM structure into the generative adversarial framework, employing the LSTM as the generator \( G \). Next, we integrate an algorithm trained on human experience into the framework by slightly modifying the structure, as illustrated in 
\begin{figure}[h!]
\centering
{\includegraphics[width=0.98\textwidth]{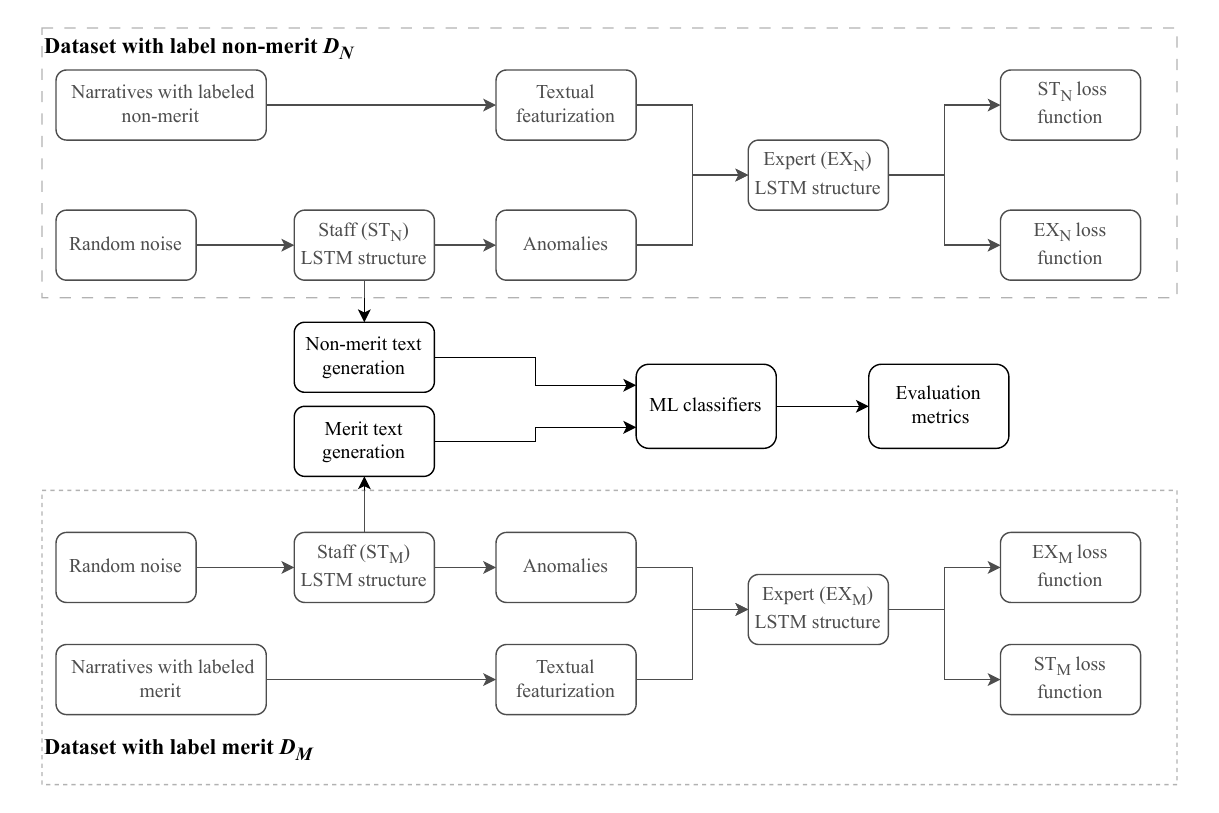}}
\caption{Generative adversarial with human-experience-trained framework.}
\label{GAN_narr}
\end{figure}
Figure~\ref{GAN_narr}.  

The proposed methodology is divided into two distinct segments. The upper segment is dedicated to generating text from narratives classified as non-meritorious, while the lower segment focuses on narratives classified as meritorious. Evaluation metrics for each subgroup are computed as described in the preceding section.  
Compared to the original GAN architecture, we introduce additional noise into the staff-trained LSTM module. In real-world scenarios, data are rarely perfect and often contains inherent noise. Adding noise to the generated fake samples enables the LSTM module to accommodate these random variations better, thereby enhancing the  performance of model on real data. Furthermore, many real-world systems invariably include uncontrollable stochastic factors. By injecting noise into the fake data, we can partially emulate these influences, aligning the operation of the entire GAN system more closely with actual systems. 
The primary objective for Figure~\ref{GAN_narr} is to develop a generative model that produces accurately labeled textual data for both categories, while maintaining performance metrics comparable to those of the original dataset.

By processing the structure presented in Figure~\ref{GAN_narr}, we generated a dataset of 5,000 textual entries, each comprising approximately 50 tokens per sentence, with labels assigned as either meritorious or non-meritorious (5,000 rows for each label). Next, we apply the evaluation metrics defined in Section~\ref{Eva_met}, and assess the ability of model to predict the meritorious labels for the generated textual dataset. Subsequently, we compare the performance between the generated dataset and the actual dataset to demonstrate that our generative model successfully replicates the patterns observed in the \citet{CFPB}.
\begin{table}[h!]
\centering
\caption{Performance summary in the percentage of ML classifiers on simulated data with Figure~\ref{GAN_narr} framework via FinBERT featurization approach.}
\begin{tabular}{crrrr}
\hline\hline 
Evaluation metrics & LR & RF & LSTM & FinBERT \\ \hline
$\Acc$ & 52.13 & 54.32 & 72.31 & 77.35 \\ 
$\Fone$ score & 51.47 & 53.88 & 72.10 & 76.88 \\ 
$\MCC$ & 18.13 & 17.95 & 49.83 & 50.09 \\ 
$\mathrm{C}_{\kappa}$ & 18.33 & 17.59 & 49.74 & 49.97 \\ \hline
\end{tabular}
\label{table1-sim}
\end{table}

Building upon the framework presented in Table~\ref{table1-true}, which demonstrates that the FinBERT featurization approach achieves superior performance when integrated with the deep model, Table~\ref{table1-sim} provides a corresponding analysis using the simulated data illustrated in Figure~\ref{GAN_narr}. Deep models such as LSTM and FinBERT exhibit notably superior performance across four distinct evaluation metrics. Nevertheless, compared to the results from the original dataset, there remains a slight difference in label accuracy, although the overall performance is remarkably similar. Overall, a comparison between Table~\ref{table1-true} and Table~\ref{table1-sim} reveals a high degree of similarity between the results, indicating that the simulated data effectively replicates key relationships observed in the actual dataset. These findings suggest that simulated data can serve as a reliable proxy when access to the actual dataset is limited.

\section{Conclusion}
\label{Conclued}
In this research, we demonstrate that expert labeling consistently outperforms staff labeling, as evidenced by significant improvements across multiple evaluation metrics, including $\Acc$, $\Fone$ score, $\MCC$, and $\mathrm{C}_{\kappa}$. Hence, we conclude that implementing expert labels is a more effective strategy for achieving higher accuracy and reliability. Furthermore, our findings underscore the importance of leveraging domain expertise to enhance the overall quality of the labeling process.

Based on expert annotations, we subsequently applied this knowledge to summarize our textual dataset for a subsequent text classification task. The original and the summarized texts were input into ML models for classification, and the results indicated comparable performance. Therefore, the summarized text is advantageous due to its reduced word space, which enhances computational efficiency. Furthermore, we implemented three distinct word embedding methods to assess the benefits of each approach, with our final results demonstrating that the FinBERT model achieved superior performance. Finally, by integrating a GANs-based model, we successfully generated simulated datasets that exhibit patterns analogous to real-world data. This approach mitigates the challenges of acquiring high-quality, accurate data (e.g., high costs, privacy concerns, and inconsistent data quality) while providing a controlled and reproducible experimental environment. Collectively, these results advocate for a hybrid strategy that leverages expert annotations, deep learning architectures, and synthetic data generation to advance the field of text classification.

\subsection{Future works}
Our current results demonstrate excellent performance in text classification and generation tasks using the FinBERT pre-trained model. FinBERT is built on the same transformer architecture introduced by BERT from Google, but it has been further pre-trained on a large corpus of financial texts. This specialized training enables FinBERT to understand and analyze financial language better. In the next phase, we intend to develop our pre-trained model based on the BERT architecture by incorporating additional complaint and insurance data. We anticipate this approach will yield more significant performance improvements in future deep-model analysis.

Regarding text generation, GAN-based experience has proven to be a highly valuable model for simulating synthetic data; however, it may not always produce the most faithful representations of the original complaint data. Therefore, exploring and implementing additional text simulation methods is essential to generate synthetic data that more accurately reflects the nuances and complexities of the original dataset. Methods such as variational autoencoders and diffusion models offer promising alternatives for future work. Variational autoencoders learn a continuous latent representation of complaint narratives and can generate diverse text by sampling from this latent space. In contrast, diffusion models iteratively refine random noise into coherent text, achieving high diversity and fidelity to the original data distribution. Retrieval-augmented generation combines retrieval techniques with generative models to condition the output on real complaint examples. This ensures the synthesized text preserves essential semantic and stylistic details crucial for robust classification.

Furthermore, the grammatical correctness of generated text likely depends on both the quality of the data and the gradient propagation properties of model when processing discrete text data. Hence, enhancing grammatical correctness by incorporating natural language generation techniques that focus on producing coherent text from structured data using linguistic rules and contextual information to improve accuracy is necessary. Contemporary natural language generation models are trained on extensive datasets to create clear and grammatically correct text.

\end{document}